%% file: main.tex
\documentclass[10pt,twocolumn,letterpaper]{article}

\usepackage{cvpr}
\usepackage{times}
\usepackage{epsfig}
\usepackage{amsmath}
\usepackage{amssymb}
\usepackage{helvet}
\usepackage{courier} 
\usepackage[hyphens]{url} 
\usepackage{float}
\usepackage{multirow}
\usepackage{graphicx} 
\usepackage{color, colortbl}
\definecolor{LightGray}{rgb}{0.92,0.92,0.92}
\usepackage[ruled,vlined]{algorithm2e}

\SetCommentSty{mycommfont}
\usepackage[title]{appendix}
\usepackage[pagebackref=true,breaklinks=true,letterpaper=true,colorlinks,bookmarks=false]{hyperref}
\usepackage{makecell}
\usepackage{xcolor} 
\usepackage{pifont}
\newcommand{\zyang}[1]{{\color[rgb]{0,0,1}{\tiny\textbf{ZY:}}{\normalsize\itshape#1}}}
\newcommand{\eat}[1]{}

\def\cvprPaperID{5007} 

\cvprfinalcopy
\ifcvprfinal\fi

\begin{document}

\title{TAP: Text-Aware Pre-training for Text-VQA and Text-Caption}

\author{Zhengyuan Yang$^1$\thanks{This work was done while Z.Yang was an intern at Microsoft.} \quad Yijuan Lu$^2$ \quad Jianfeng Wang $^2$ \quad Xi Yin$^{2}$ \\ \quad Dinei Florencio$^2$ \quad Lijuan Wang$^2$ \quad Cha Zhang$^2$ \quad Lei Zhang$^2$ \quad Jiebo Luo$^1$\\ $^1$University of Rochester \qquad\qquad $^2$Microsoft Corporation \\ 
}

\maketitle

\begin{abstract}
  \input{abs}
\end{abstract}

\input{intro}
\input{related}
\input{approach}
\input{exp}
\input{conclusion}

\clearpage
{\small
\section*{Acknowledgment}
Zhengyuan Yang and Jiebo Luo were supported in part by NSF awards IIS-1704337, IIS-1722847, and IIS-1813709.
\bibliographystyle{ieee_fullname}
\bibliography{ref.bib}
}
\clearpage
\appendix
\input{supply}
\end{document}

%% file: abs.tex
In this paper, we propose Text-Aware Pre-training (TAP) for Text-VQA and Text-Caption tasks. These two tasks aim at reading and understanding scene text in images for question answering and image caption generation, respectively. In contrast to conventional vision-language pre-training that fails to capture scene text and its relationship with the visual and text modalities, TAP explicitly incorporates scene text (generated from OCR engines) during pre-training. With three pre-training tasks, including masked language modeling (MLM), image-text (contrastive) matching (ITM), and relative (spatial) position prediction (RPP), pre-training with scene text effectively helps the model learn a better aligned representation among the three modalities: text word, visual object, and scene text. Due to this aligned representation learning, even pre-trained on the same downstream task dataset, TAP already boosts the absolute accuracy on the TextVQA dataset by $+5.4\%$, compared with a non-TAP baseline. To further improve the performance, we build a large-scale scene text-related image-text dataset based on the Conceptual Caption dataset, named OCR-CC, which contains $1.4$ million images with scene text. Pre-trained on this OCR-CC dataset, our approach outperforms the state of the art by large margins on multiple tasks, \ie, $+8.3\%$ accuracy on TextVQA, $+8.6\%$ accuracy on ST-VQA, and $+10.2$ CIDEr score on TextCaps.

%% file: intro.tex
\section{Introduction}
\begin{figure}[t]
\begin{center}
  \centerline{\includegraphics[width=8cm]{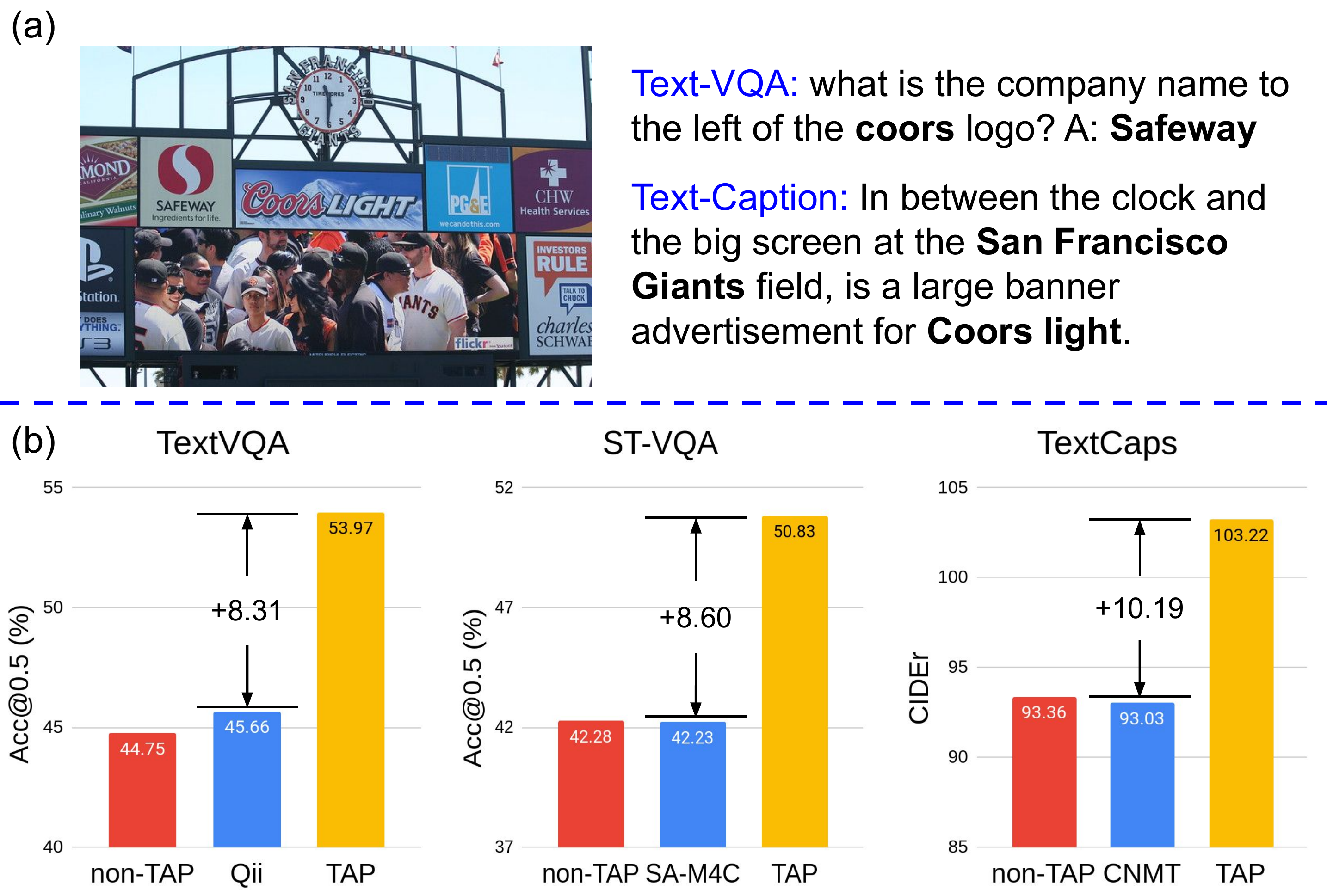}}
\end{center}
\vspace{-0.2in}
    \caption{\textbf{(a)} Text-VQA and Text-Caption tasks aim at reading and understanding scene text in images for question answering and image caption generation, respectively. We highlight the scene text-related words in bold. \textbf{(b)} By explicitly incorporating scene text in pre-training, Text-Aware Pre-training (TAP) significantly outperforms both the non-TAP baseline and previous state of the art on multiple tasks (bars shown in red and blue colors, respectively).
	}
\label{fig:intro}
\end{figure}

\vspace{-1pt}
The \textit{Vision-language tasks incorporating scene text}~\cite{bigham2010vizwiz,gurari2018vizwiz,singh2019towards,sidorov2020textcaps}, \eg, Text-VQA~\cite{singh2019towards,biten2019scene,mishra2019ocr,wang2020general} and Text-Caption~\cite{sidorov2020textcaps}, pose new challenges to vision-language models of reading and understanding scene text in image context. 
Extended from Visual Question Answering (VQA)~\cite{VQA_15}, Text-VQA aims to answer questions by understanding the scene text in the image-question context. Text-Caption seeks to generate an image caption~\cite{veit2016coco,anderson2018bottom} that describes both the visual and scene text information in the image, as shown in Figure~\ref{fig:intro} (a). These tasks have many potential applications, including robotics~\cite{anderson2018vision}, document understanding~\cite{mishra2019ocr}, assisting visually-impaired people~\cite{bigham2010vizwiz,gurari2018vizwiz}, \etc.

A typical Text-VQA/Text-Caption framework consists of 1) a feature encoder for each single modality (text word, visual object, and scene text), 2) a multi-modal fusion module, and 3) a decoding module for prediction generation. Previous studies~\cite{singh2019towards,gao2020multi,gao2020structured,hu2020iterative,kant2020spatially,sidorov2020textcaps,wang2020multimodal} improve the model's performance by designing stronger network architectures. Among them, LoRRA~\cite{singh2019towards} added an OCR attention branch for scene text encoding to a VQA model~\cite{jiang2018pythia}. M4C~\cite{hu2020iterative,sidorov2020textcaps} proposed a transformer-based multi-modal fusion module~\cite{vaswani2017attention} and a multi-step multi-choice decoding module.
Despite the effective network design, most previous models are optimized with a sole objective directly towards the correct answer/caption.
Such a single answer/caption loss tries to predict each word in the ground-truth but is less effective in learning a joint representation among text word, visual object, and scene text.
Without a good joint representation, directly optimizing for question-answering/image-captioning could be challenging.
Inspired by the success of Vision-Language Pre-training (VLP)~\cite{lu2019vilbert,li2019visualbert,chen2019uniter,tan2019lxmert,li2020oscar,huang2020pixel,cao2020behind} in image-text joint representation learning, we leverage the effective Text-VQA/Text-Caption network designs and explore to further improve Text-VQA/Text-Caption by pre-training.

Vision-Language Pre-training (VLP) shows its effectiveness in learning task-agnostic joint representations of image and text. The main idea is to first pre-train the model with pre-training tasks on image-caption datasets~\cite{sharma2018conceptual,krishna2017visual,veit2016coco,ordonez2011im2text,plummer2015flickr30k}, and then fine-tune the model for a specific vision-language task~\cite{VQA_15,young2014image,kazemzadeh2014referitgame,veit2016coco}. However, conventional VLP methods are designed intuitively for vision-language tasks and do not include scene text in pre-training. Therefore, previous methods fail to capture the scene text modality and its relationship with the visual and text modalities, and are thus less effective in Text-VQA/Text-Caption. 

In this study, we propose \textit{Text-Aware Pre-training} (TAP), which incorporates the scene text modality in pre-training to learn a joint representation of text word, visual object, and scene text. In TAP, we design text-aware pre-training tasks to better fuse scene text (including both scene text words and their visual regions detected by OCR) with the \textit{text words} and \textit{visual objects}. For the former, we refine the pre-training tasks in VLP~\cite{lu2019vilbert,li2020oscar} to support the extra scene text input. We find it particularly important to include the detected scene text words as extra language inputs. The extra inputs anchor the scene text and language modalities and make the aligned representation learning easier.
For the latter, previous studies~\cite{kant2020spatially,wang2020multimodal} show that the spatial relationships between scene text and object regions are important, \eg, the relationship ``left'' in Figure~\ref{fig:intro} (a). Therefore, we propose a ``relative (spatial) position prediction'' task that learns regions' spatial relationships by predicting their relative spatial positions in pre-training.

The extra scene text modality, together with the specially designed pre-training tasks, effectively helps the model learn a better aligned representation among the three modalities: text word, visual object, and scene text. This aligned representation learning, even pre-trained and fine-tuned on the same downstream task dataset\eat{ with less than $30$K images}, leads to significant improvement over the non-TAP baseline and helps the TAP model achieve the new state of the art.

To further unleash the power of TAP, we clean and generate a large-scale scene text-related image-caption dataset for pre-training. In general image-caption datasets~\cite{sharma2018conceptual,krishna2017visual,veit2016coco,ordonez2011im2text,plummer2015flickr30k}, many image-text pairs contain either no scene text-related visual regions or no scene text-related language referring, and are thus less helpful to Text-VQA/Text-Caption. On the visual side, we run an OCR detector to filter out images with no scene text. On the language side, we include the detected OCR text tokens as the additional caption input to obtain scene text-related language descriptions. In the end, we build a large-scale dataset named OCR-CC with around $1.4$ million scene text-related image-text pairs based on the Conceptual Captioning dataset~\cite{sharma2018conceptual}. By using this large-scale dataset for pre-training, we observe further improvement on the Text-VQA and Text-Caption tasks.

We experiment with the TAP approach on the M4C network architecture~\cite{hu2020iterative} and benchmark it on the TextVQA~\cite{singh2019towards}, ST-VQA~\cite{biten2019scene}, and TextCaps~\cite{sidorov2020textcaps} datasets. With the identical network architecture and training data, TAP improves the accuracy on the TextVQA dataset~\cite{singh2019towards} from $44.50\%$ to $49.91\%$, compared with a non-TAP baseline. Our final model ranks \textbf{No.1}\footnote{According to the official leader-boards (Nov. 2020)} on multiple Text-VQA/Text-Caption challenges, and outperforms previous methods by large margins: TextVQA~\cite{singh2019towards} ($+8.3\%$ in absolute accuracy), ST-VQA~\cite{biten2019scene} ($+8.6\%$ in absolute accuracy), and TextCaps~\cite{sidorov2020textcaps} ($+10.2$ in CIDEr score).

Our main contributions are:
\vspace{-6pt}
\begin{itemize} 
\setlength\itemsep{-3pt}
\item To the best of our knowledge, we are the first to explore pre-training for Text-VQA and Text-Caption.
\item By explicitly incorporating scene text with three specially designed pre-training tasks, Text-Aware Pre-training (TAP) effectively learns a better aligned representation that leads to significant performance improvement on Text-VQA/Text-Caption.
\item We build a large-scale dataset named OCR-CC with around $1.4$ million scene text-related image-text pairs. TAP with OCR-CC leads to the new state of the art on multiple tasks: TextVQA~\cite{singh2019towards} ($+8.3\%$ in absolute accuracy), ST-VQA~\cite{biten2019scene} ($+8.6\%$ in absolute accuracy), and TextCaps~\cite{sidorov2020textcaps} ($+10.2$ in CIDEr score).
We will release the dataset and the models.
\vspace{-3pt}
\end{itemize}

%% file: related.tex
\section{Related Work}
\begin{figure*}[t]
\vspace{-11pt}
\begin{center}
  \centerline{\includegraphics[width=17cm]{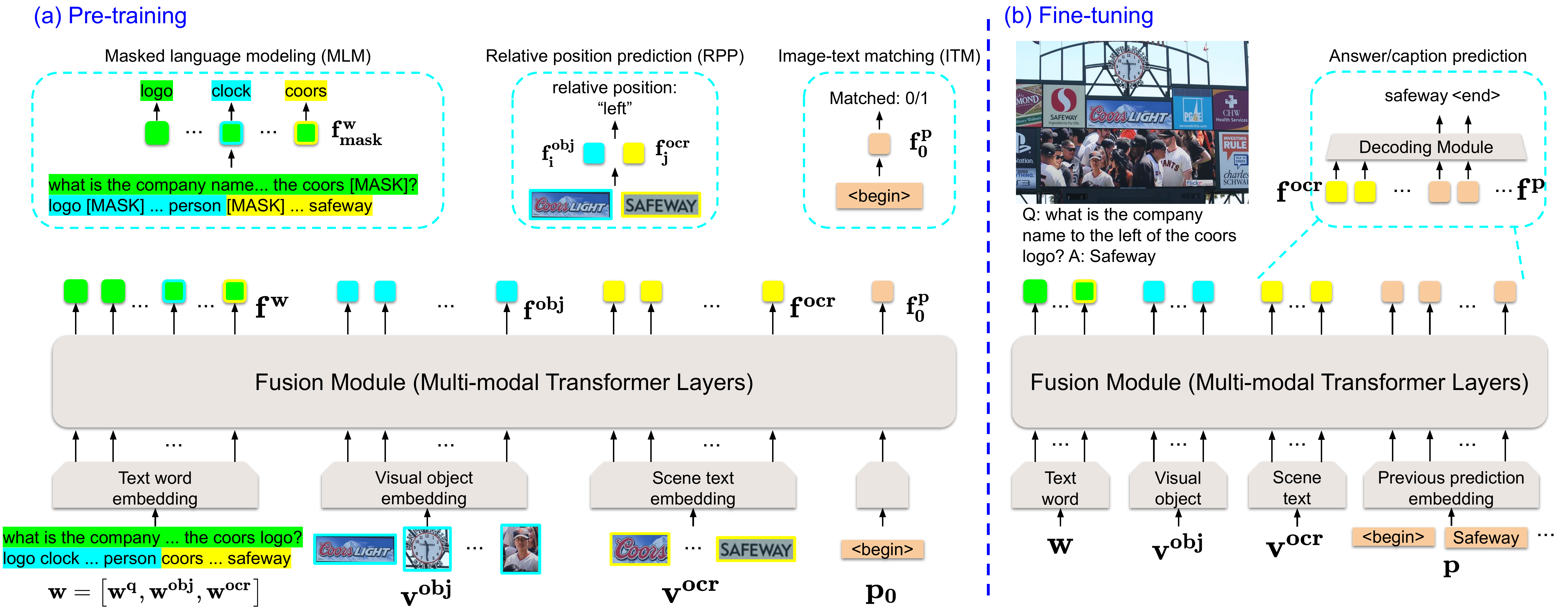}}
\end{center}
\vspace{-0.25in}
	\caption{
	An overview of Text-Aware Pre-training (TAP). \textbf{(a)} In pre-training, the framework takes text words $\mathbf{w}$, visual objects $\mathbf{v^{obj}}$, scene text $\mathbf{v^{ocr}}$, and a special $\texttt{begin}$ token $\mathbf{p_0}$ as inputs, and improves the aligned representation learning by performing pre-training tasks (MLM, ITM, RPP) on fused feature $\mathbf{f}$. \textbf{(b)} In fine-tuning, we train the same model to step-by-step generate the answer/caption prediction, conditioned on $\mathbf{w}$, $\mathbf{v^{obj}}$, $\mathbf{v^{ocr}}$, and the previous word predictions $\mathbf{p}=\{\mathbf{p_t}\}_{t=0}^{T-1}$ at decoding step $T$. Text word, visual object, and scene text-related tokens are highlighted by the green, cyan, and yellow colors, respectively.}
\label{fig:arch}
\vspace{-0.15in}
\end{figure*}

\noindent\textbf{Vision-language tasks incorporating scene text.}
Text-VQA~\cite{singh2019towards,biten2019scene,mishra2019ocr,wang2020general} and Text-Caption~\cite{sidorov2020textcaps} aim at reading and understanding scene text in images for question answering and image caption generation. Various datasets~\cite{singh2019towards,biten2019scene,mishra2019ocr} are built for the Text-VQA task, \eg, the TextVQA dataset~\cite{singh2019towards}, the ST-VQA dataset~\cite{biten2019scene}, \etc. TextCaps~\cite{sidorov2020textcaps} is a dataset recently proposed for the Text-Caption task.

Recent studies~\cite{singh2019towards,gao2020multi,gao2020structured,hu2020iterative,kant2020spatially,wang2020multimodal,liu2020cascade,han2020finding} proposed various network architectures to improve the Text-VQA/Text-Caption performance. Among them, LoRRA~\cite{singh2019towards} approached Text-VQA by extending a VQA model Pythia~\cite{jiang2018pythia} with an OCR attention branch. The answer vocabulary is a combination of a static vocabulary and detected OCR tokens. Multi-modal Multi-Copy Mesh (M4C)~\cite{hu2020iterative} boosted the Text-VQA performance by proposing a transformer-based multi-modal fusion module~\cite{vaswani2017attention} and a multi-step multi-choice decoding module that supports multi-step answer decoding. M4C's variants M4C-Captioner~\cite{sidorov2020textcaps} set a strong baseline on TextCaps~\cite{sidorov2020textcaps} with the question text inputs removed. SA-M4C~\cite{kant2020spatially} further improved M4C by encoding the spatial relationships among visual regions as the attention masks in the multi-modal transformer. Similar explorations~\cite{wang2020multimodal} on the spatial relationships are studied in the Text-Caption task.

Despite the effective network design, all previous studies directly optimize towards the sole objective for the Text-VQA/Text-Caption task. We contend that such a single answer/caption loss could be ineffective in aligned representation learning and thus limits the Text-VQA/Text-Caption performance. In this study, we leverage the effective network designs and explore to further improve Text-VQA/Text-Caption by pre-training.

\noindent\textbf{Vision-Language Pre-training (VLP).}
VLP~\cite{lu2019vilbert,li2019visualbert,alberti2019fusion,li2020unicoder,tan2019lxmert,su2019vl,zhou2020unified,chen2019uniter,lu202012,li2020oscar,huang2020pixel} shows its effectiveness in learning task-agnostic vision-language joint representations.
Most studies~\cite{lu2019vilbert,tan2019lxmert,chen2019uniter} focused on vision-language understanding tasks, \eg, image-text retrieval~\cite{young2014image}, visual question answering~\cite{VQA_15}, visual grounding~\cite{kazemzadeh2014referitgame}, \etc. Recent studies~\cite{zhou2020unified,li2020oscar,hu2020vivo} unified the pre-training framework to cover generation tasks, \eg, image captioning~\cite{veit2016coco,anderson2018bottom}.

However, conventional VLP methods do not capture scene text during pre-training and are therefore less effective for Text-VQA/Text-Caption. The proposed Text-aware Pre-training (TAP) explicitly incorporates scene text to learn a better aligned representation among the three modalities: text word, visual object, and scene text.

%% file: approach.tex
\section{Text-Aware Pre-training (TAP)}

TAP explicitly incorporates scene text in pre-training to improve Text-VQA/Text-Caption. We first pre-train the model with the scene text-aware pre-training tasks and then fine-tune it for a specific downstream task. 

In this section, we first introduce the design of scene text-aware pre-training tasks. We then present the data corpus used for TAP and our proposed OCR-CC dataset. We postpone the model details to Section~\ref{sec:setting}.

\subsection{Text-aware pre-training tasks}
Figure~\ref{fig:arch} overviews TAP in pre-training and fine-tuning. In pre-training, the input to the fusion module are embeddings of ${K}$ text words $\mathbf{w}$, ${M}$ object regions $\mathbf{v^{obj}}$, ${N}$ scene text regions $\mathbf{v^{ocr}}$, and a special $\texttt{begin}$ token $\mathbf{p_0}$. In the text word embedding, each word in the extended text input $\mathbf{w} = \left[\mathbf{w^q}, \mathbf{w^{obj}}, \mathbf{w^{ocr}}\right]$ is encoded as a feature vector, where $\mathbf{w^q}, \mathbf{w^{obj}}, \mathbf{w^{ocr}}$ are the question text, detected object labels, and detected scene text words. In the object and scene text embedding, object and scene text regions are detected and encoded by object detectors and OCR engines. 

Taking the fused feature $\mathbf{f}=\left[\mathbf{f^w}, \mathbf{f^{obj}}, \mathbf{f^{ocr}}, \mathbf{f^p}\right]$ as inputs, TAP improves multi-modal fusion by performing text-aware pre-training tasks. The proposed pre-training tasks consist of two parts, focusing on fusing scene text $\mathbf{v^{ocr}}$ with text words $\mathbf{w}$ and visual objects $\mathbf{v^{obj}}$, respectively.

\noindent\textbf{Scene-text language pre-training tasks.}
To better fuse the scene text $\mathbf{v^{ocr}}$ with the text words $\mathbf{w}$, we design two scene-text language pre-training tasks based on the masked language modeling (MLM) and image-text (contrastive) matching (ITM) tasks in VLP~\cite{devlin2018bert,lu2019vilbert,chen2019uniter}.
For MLM on the extended text input $\mathbf{w} = \left[\mathbf{w^q}, \mathbf{w^{obj}}, \mathbf{w^{ocr}}\right]$, we randomly mask each text token in $\mathbf{w}$ with a probability of $15\%$. The masked words $\mathbf{w_{mask}}$ are replaced with a special $\texttt{MASK}$ token $80\%$ of the time, a random word $10\%$, and remains unchanged $10\%$. The MLM task takes the fused feature at the masked position $\mathbf{f_{mask}^w}$ as the input, and aims to recover the masked word $\mathbf{w_{mask}}$ with two fully-connected layers. For ITM, $\mathbf{w}$ is polluted $50\%$ of the time by replacing text sub-sequence $\mathbf{w^q}$, $\mathbf{w^{obj}}$, or $\mathbf{w^{ocr}}$ with a randomly-selected one from another image. The polluted text words $\mathbf{w}$ are thus not paired with the visual regions $\mathbf{v^{obj}}$ and $\mathbf{v^{ocr}}$. The ITM task takes the sequence feature $\mathbf{f_{0}^p}$ as the input and aims to predict if the sequence has been polluted or not.

We find that the extra scene text word input $\mathbf{w^{ocr}}$ is critical for learning the scene-text language aligned representation.
As a comparison to the extended text input $\mathbf{w}$, pre-training with the original MLM and ITM~\cite{devlin2018bert,lu2019vilbert} on question text $\mathbf{w^q}$ leads to limited improvement over the non-pre-training baseline. The failure is due to the limited number of scene text-related words in the language input $\mathbf{w^q}$. In this case, since many randomly masked words $\mathbf{w^q_{mask}}$ and polluted sequences are not relevant to scene text, scene text regions $\mathbf{v^{ocr}}$ are less important for solving the pre-training tasks (MLM, ITM) and are thus often overlooked. $\mathbf{w^{ocr}}$ in the extended text input $\mathbf{w}$ generates extra scene text referring in the language modality and thus makes TAP effective.

\noindent\textbf{Scene-text visual pre-training tasks.}
Understanding the spatial relationships between the visual object $\mathbf{v^{obj}}$ and scene text $\mathbf{v^{ocr}}$ benefits Text-VQA/Text-Caption~\cite{kant2020spatially,wang2020multimodal}.
The extra feature input of bounding box coordinates helps the spatial relationship learning~\cite{hu2020iterative,gao2020multi,gao2020structured}, but hasn't fully solved the problem.
Recent studies~\cite{kant2020spatially,wang2020multimodal} hard code the coordinate features as the regions' relationships in feature fusion and obtain further improvement. In this study, we explore spatial relationship learning by pre-training.

Specifically, we design a scene-text visual pre-training task in TAP. The main idea is to predict the relative spatial position between two randomly sampled visual regions. Therefore, we refer to the task as ``relative (spatial) position prediction'' (RPP). The input to the pre-training task is a randomly sampled visual object feature $\mathbf{f_i^{obj}}$ and scene text feature $\mathbf{f_j^{ocr}}$, where $i \in \{1,\cdots, M\}$ and $j\in \{1,\cdots, N\}$. The objective is to predict the relative spatial position between the two sampled regions $\mathbf{v_i^{obj}}$ and $\mathbf{v_j^{ocr}}$. We start with a single relationship of whether ``scene text region $\mathbf{v_j^{ocr}}$ is \textit{on} object $\mathbf{v_i^{obj}}$,'' and thus model RPP as a binary classification problem. We then extend the task to a 12-class relative position prediction problem with the classes defined by Yao~\etal~\cite{yao2018exploring}, including on, cover, overlap, eight-way relative orientation, and unrelated. 

\subsection{Pre-training corpus}
\label{sec:train}
TAP works well even without extra pre-training data. We first experiment with ``TAP without extra data,'' where we only use the downstream Text-VQA/Text-Caption dataset for pre-training, \ie, the training set of the TextVQA~\cite{singh2019towards}, ST-VQA~\cite{biten2019scene}, or TextCaps~\cite{sidorov2020textcaps} datasets. These datasets~\cite{singh2019towards,biten2019scene,sidorov2020textcaps} all contain less than $30$K images and $150$K image-text pairs. We detail the pre-training and fine-tuning pipeline for each downstream task in Section~\ref{sec:setting}.

We then experiment with ``TAP with large-scale data.'' We build a large-scale scene text-related image-caption dataset named \textit{OCR-CC} based on the Conceptual Caption (CC) dataset~\cite{sharma2018conceptual}, and use the dataset for pre-training. 
Among the image-caption datasets~\cite{sharma2018conceptual,krishna2017visual,veit2016coco,ordonez2011im2text,plummer2015flickr30k}, only the CC dataset contains a reasonable portion of images with meaningful scene text regions. Therefore, we run the Microsoft Azure OCR system\footnote{Public Microsoft OCR API\eat{ (June 2020)}: \url{https://docs.microsoft.com/en-us/azure/cognitive-services/computer-vision/concept-recognizing-text}} on all images in the CC dataset and filter out the images with no scene text, watermarks only, and tiny scene text regions only. In the end, we obtain $1.367$ million image-caption pairs with a mean and median of $11.4$ and $6$ scene text detected per image. As a reference, the mean and median are $23.1$ and $12$ in the TextVQA dataset~\cite{hu2020iterative}, and $8.03$ and $6$ in the ST-VQA dataset~\cite{biten2019scene}. We adopt the same region feature extraction method used in the TextVQA dataset~\cite{singh2019towards} to provide object and scene text region embedding. By including scene text words $\mathbf{w^{ocr}}$ as additional text inputs, OCR-CC provides scene text-related image-caption pairs for TAP. We keep the caption text from CC in OCR-CC and use it as the question text $\mathbf{w^q}$ in pre-training.
We show the details of dataset collection, scene text number distribution, and additional qualitative examples of OCR-CC in the supplementary material.

%% file: exp.tex
\section{Experiments}
We benchmark TAP for both the Text-VQA task on the TextVQA~\cite{singh2019towards} and ST-VQA~\cite{biten2019scene} datasets, and the Text-Caption task on the TextCaps dataset~\cite{sidorov2020textcaps}. We use our proposed OCR-CC dataset for large-scale pre-training.

\subsection{Datasets}
\label{sec:dataset}

\noindent\textbf{TextVQA.}
The TextVQA dataset~\cite{singh2019towards} contains 28,408 images from the Open Images dataset~\cite{kuznetsova2018open}. We follow the same training/validation/test split used in the previous work~\cite{singh2019towards} in our experiments. The methods are evaluated by the soft-voting accuracy of 10 answers. 

\noindent\textbf{ST-VQA.}
The ST-VQA dataset~\cite{biten2019scene} contains 21,892 images from multiple sources including ICDAR 2013~\cite{karatzas2013icdar}, ICDAR 2015~\cite{karatzas2015icdar}, ImageNet~\cite{deng2009imagenet}, VizWiz~\cite{gurari2018vizwiz}, IIIT STR~\cite{mishra2013image}, Visual Genome~\cite{krishna2017visual}, and COCO-Text~\cite{veit2016coco}. The methods are evaluated by both accuracy and Average Normalized Levenshtein Similarity (ANLS)~\cite{biten2019scene}.

\noindent\textbf{TextCaps.}
The TextCaps dataset~\cite{sidorov2020textcaps} augments the 28,408 images in TextVQA~\cite{singh2019towards} with 145,329 captions. The captions are evaluated by the caption metrics (BLEU~\cite{papineni2002bleu}, METEOR~\cite{denkowski2014meteor}, ROUGE\_L~\cite{lin2004rouge}, SPICE~\cite{anderson2016spice}, and CIDEr~\cite{vedantam2015cider}).

\noindent\textbf{OCR-CC.}
Our OCR-CC dataset contains $1.367$ million scene text-related image-caption pairs from the Conceptual Captioning (CC) dataset~\cite{sharma2018conceptual}. More details of OCR-CC are in the supplementary material.

\subsection{Experiment settings}
\label{sec:setting}
\noindent\textbf{Network architecture.}
We conduct experiments based on the M4C network architecture~\cite{hu2020iterative}. We extend the text input $\mathbf{w_q}$ with the object labels $\mathbf{w^{obj}}$ and scene text words $\mathbf{w^{ocr}}$. We keep all remaining settings the same as in the original M4C~\cite{hu2020iterative}, including the feature embedding, network architecture, training parameters, and layer initialization. 

M4C's text encoder is a three-layer trainable transformer~\cite{vaswani2017attention} initialized from the first three layers of BERT$_\text{BASE}$~\cite{devlin2018bert}. A pre-trained Faster R-CNN~\cite{ren2015faster} detects objects and represents the detected region with its visual and coordinate features. The final layer (fc7) of the detector is fine-tuned. An offline OCR detector~\cite{borisyuk2018rosetta} detects scene text regions and represents the region with its visual, coordinates, FastText~\cite{bojanowski2017enriching}, and Pyramidal Histogram of Characters (PHOC)~\cite{almazan2014word} features. The fusion module in M4C is a four-layer multi-modal transformer that has the same hyper-parameters as BERT$_\text{BASE}$. The fusion module is initialized from scratch. A multi-step decoding module then takes fused features $\mathbf{f^{ocr}},\mathbf{f^{p}}$ as inputs, and word-by-word predicts the final answer. The predicted answer word at each decoding step $T$ is selected either from a fixed frequent word vocabulary or from the dynamic OCR tokens. The word classification loss is applied to each decoding step. 

\noindent\textbf{Adapting to Text-VQA.}
By taking the fused feature $\mathbf{f}$ as input, we pre-train the feature encoder and fusion module with the pre-training tasks (MLM, ITM, RPP). MLM is only computed on the sequences that have not been polluted by ITM. The pre-trained model with the highest pre-training task accuracy is used to initialize the feature encoder and fusion module. In fine-tuning, the model step-by-step predicts the answer with an extra decoding module,  and is trained with the answer classification loss in each step.

\noindent\textbf{Adapting to Text-Caption.}
We keep the framework architecture the same for Text-Caption as for Text-VQA, except increasing the maximum answer decoding length from $12$ words~\cite{hu2020iterative} to $30$ words~\cite{sidorov2020textcaps}. $\mathbf{w^q}$ is left blank in both pre-training and fine-tuning. The input text sequence $\mathbf{w}$ consists of $\mathbf{w^{ocr}}$, $\mathbf{w^{obj}}$, and the blank $\mathbf{w^q}$. During fine-tuning, the framework is trained with the same multi-step word classification loss as used in Text-VQA. 

\noindent\textbf{Compared methods.}
We compare TAP with other state of the art~\cite{singh2019towards,gao2020multi,hu2020iterative,kant2020spatially,gao2020structured,liu2020cascade,han2020finding,wang2020multimodal} and systematically study the following baselines and variants of our method.

\vspace{-9pt}
\begin{itemize} 
\setlength\itemsep{-3pt}
\item \textbf{TAP~(Ours).} We first experiment with ``TAP without extra pre-training data.'' We use the same downstream task dataset for both pre-training and fine-tuning, and follow the same training parameters as used in M4C. For the Text-VQA task, we pre-train the model for $24$K iterations with the pre-training tasks (MLM, ITM, RPP) and then fine-tune it with the answer loss for another $24$K iterations. The numbers of pre-training and fine-tuning iterations are both $12$K for the Text-Caption task following M4C-Captioner~\cite{sidorov2020textcaps}. 
\item \textbf{M4C$^\dagger$.} ``M4C$^\dagger$'' is the non-TAP baseline. Based on M4C, we include the detected object labels $\mathbf{w^{obj}}$ and scene text tokens $\mathbf{w^{ocr}}$ as the additional text input following ``TAP.'' We train the model for $48$K iterations with the answer loss to match TAP's total iteration number. Compared with ``TAP,'' the only difference is that ``M4C$^\dagger$'' trains the first $24$K iterations with the answer loss, instead of the pre-training tasks.
\item \textbf{TAP$^{\dagger\dagger}$~(Ours).} ``TAP$^{\dagger\dagger}$'' reports our best performance achieved with extra pre-training data (TextVQA, ST-VQA, TextCaps, OCR-CC) and other minor modifications. We pre-train ``TAP$^{\dagger\dagger}$'' for $480$K iterations. Section~\ref{sec:ablation} details the benefits of each extra data source.
\end{itemize}

\subsection{Text-VQA/Text-Caption results}
\label{sec:results}
\begin{table*}[t]
\centering
\caption{Text-VQA results on the TextVQA dataset~\cite{singh2019towards}. The top part reports results in the constrained setting that only uses TextVQA for training and Rosetta for OCR detection. \eat{ do not use extra training data or strengthened OCR predictions.}The bottom part compares our best performance with other state-of-the-art methods in the unconstrained setting. The methods ``M4C$^\dagger$,'' ``TAP,'' ``TAP$^{\dagger\dagger}$'' are detailed in Section~\ref{sec:setting}.}
\vspace{-0.0in}
\begin{tabular}{ l l l c c }
    \hline
    Method & OCR System & Extra Data & Val Acc. & Test Acc.\\
    \hline
    LoRRA~\cite{singh2019towards} & Rosetta-ml & \ding{55} & 26.56 & 27.63 \\
    MM-GNN~\cite{gao2020multi} & Rosetta-ml & \ding{55} & 31.44 & 31.10 \\
    M4C~\cite{hu2020iterative} & Rosetta-en & \ding{55} & 39.40 & 39.01 \\
    SMA~\cite{gao2020structured} & Rosetta-en & \ding{55} & 40.05 & 40.66 \\
    CRN~\cite{liu2020cascade} & Rosetta-en & \ding{55} & 40.39 & 40.96 \\
    LaAP-Net~\cite{han2020finding} & Rosetta-en & \ding{55} & 40.68 & 40.54 \\
    M4C$^\dagger$~\cite{hu2020iterative} & Rosetta-en & \ding{55} & 39.55 & - \\
    \rowcolor{LightGray}
    TAP~(Ours) & Rosetta-en & \ding{55} & \eat{\zyang{44.06}}44.06 & - \\
    \hline
    M4C~\cite{hu2020iterative} & Rosetta-en & ST-VQA & 40.55 & 40.46 \\
    LaAP-Net~\cite{han2020finding} & Rosetta-en & ST-VQA & 41.02 & 40.54 \\
    SA-M4C~\cite{kant2020spatially} & Google-OCR & ST-VQA & 45.4 & 44.6 \\
    SMA~\cite{gao2020structured} & SBD-Trans OCR & ST-VQA & - & 45.51 \\
    M4C$^\dagger$~\cite{hu2020iterative} & Microsoft-OCR & \ding{55} & 44.50 & 44.75 \\
    M4C$^\dagger$~\cite{hu2020iterative} & Microsoft-OCR & ST-VQA & \eat{\zyang{45.22}}45.22 & - \\
    \rowcolor{LightGray}
    TAP~(Ours) & Microsoft-OCR & \ding{55} & 49.91 & 49.71 \\
    \rowcolor{LightGray}
    TAP~(Ours) & Microsoft-OCR & ST-VQA & 50.57 & 50.71 \\
    \rowcolor{LightGray}
    TAP$^{\dagger\dagger}$~(Ours) & Microsoft-OCR & \small{ST-VQA, TextCaps, OCR-CC} & \textbf{54.71} & \textbf{53.97} \\
    \hline
\end{tabular}
\vspace{-0.1in}
\label{table:textvqa}
\end{table*}
\noindent\textbf{TextVQA.}
Table~\ref{table:textvqa} reports the accuracy on the TextVQA dataset~\cite{singh2019towards}. 
The \textbf{top part} of the table shows the results in the constrained setting that only uses TextVQA for training and Rosetta~\cite{borisyuk2018rosetta} for OCR detection. The \textbf{bottom} compares our best performance with the state of the art~\cite{singh2019towards,gao2020multi,hu2020iterative,kant2020spatially,gao2020structured,liu2020cascade,han2020finding,wang2020multimodal} in the unconstrained setting. 

We list the adopted OCR detector in the ``OCR system'' column.  LoRRA~\cite{singh2019towards} and M4C~\cite{hu2020iterative} adopted the Rosetta OCR system~\cite{borisyuk2018rosetta}. SA-M4C~\cite{kant2020spatially} and SMA~\cite{gao2020structured} experiment with both Rosetta and other OCR systems (Google-OCR, SBD-Trans OCR). In this study, we experiment with Rosetta and the Microsoft Azure OCR system (Microsoft-OCR). We use Microsoft-OCR to detect the single OCR words appeared in the image, \ie, each detected scene text region contains only a single word.
The ``Extra data'' column shows the used training data other than the TextVQA dataset. Previous methods~\cite{hu2020iterative,kant2020spatially,gao2020structured} \eat{find it helpful to }adopt the ST-VQA dataset for joint training. Other than ST-VQA, TAP enables the use of weak data with no ground-truth answer in pre-training, \eg, TextCaps and OCR-CC. ``TAP$^{\dagger\dagger}$'' reports the final performance with all extra datasets.

Three major observations can be made from Table~\ref{table:textvqa}:
\textbf{1)} ``TAP'' significantly outperforms the non-TAP baseline ``M4C$^\dagger$'' with the identical training data and network architecture, in both the constrained setting (top part of Table~\ref{table:textvqa}) and the unconstrained setting (bottom part). In the constrained setting, TAP improves the non-TAP baseline accuracy from $39.55\%$ to $44.06\%$. In the unconstrained setting, ``TAP'' with Microsoft-OCR obtain $5.4\%$ and $5.3\%$ absolute accuracy improvement over the corresponding non-TAP baselines ``M4C$^\dagger$'' and ``M4C$^\dagger$~+STVQA,'' respectively. The improvement achieved with the same network and training data validates the effectiveness of our pre-training approach for Text-VQA/Text-Caption.
\textbf{2)} ``TAP'' outperforms the previous state of the art~\cite{singh2019towards,gao2020multi,hu2020iterative,gao2020structured,liu2020cascade,han2020finding} by large margins, even without large-scale pre-training. 
\textbf{3)} Large-scale pre-training with the OCR-CC dataset further improves the accuracy. ``TAP$^{\dagger\dagger}$'' adopts OCR-CC in pre-training and improves the accuracy from $49.91\%$ to $54.71\%$. The improvement shows that TAP benefits from extra training data, and indicates the effectiveness of our proposed OCR-CC. 
\begin{table}[t]
\centering
\caption{Text-VQA results on the ST-VQA dataset~\cite{biten2019scene}.}
\vspace{-0.0in}
\begin{tabular}{ l c c c }
    \hline
    Method & Val Acc. & Val ANLS & Test ANLS\\
    \hline
    SAN+STR~\cite{biten2019scene} & - & - & 0.135 \\
    M4C~\cite{hu2020iterative} & 38.05 & 0.472 & 0.462 \\
    SA-M4C~\cite{kant2020spatially} & 42.23 & 0.512 & 0.504 \\
    SMA~\cite{gao2020structured} & - & - & 0.466 \\
    CRN~\cite{liu2020cascade} & - & - & 0.483 \\
    LaAP-Net~\cite{han2020finding} & 39.74 & 0.497 & 0.485 \\
    M4C$^\dagger$~\cite{hu2020iterative} & 42.28 & 0.517 & 0.517 \\
    \rowcolor{LightGray}
    TAP~(Ours) & 45.29 & 0.551 & 0.543 \\
    \rowcolor{LightGray}
    TAP$^{\dagger\dagger}$~(Ours) & \textbf{50.83} & \textbf{0.598} & \textbf{0.597} \\
    \hline
\end{tabular}
\vspace{-0.0in}
\label{table:stvqa}
\end{table}

\noindent\textbf{ST-VQA.}
Table~\ref{table:stvqa} shows the Text-VQA accuracy on the ST-VQA dataset~\cite{biten2019scene} in the unconstrained setting. 
``TAP'' uses the Microsoft-OCR and is pre-trained and fine-tuned on the training set of ST-VQA. ``TAP$^{\dagger\dagger}$'' uses TextVQA, ST-VQA, TextCaps, and OCR-CC in pre-training. Similar conclusions as in Table~\ref{table:textvqa} can be drawn from Table~\ref{table:stvqa}. First, ``TAP'' outperforms the state of the art~\cite{hu2020iterative,kant2020spatially,gao2020structured,liu2020cascade,han2020finding} by large margins, and significantly improves the non-TAP baseline ``M4C$^\dagger$.'' Second, large-scale pre-training further improves the accuracy by $+5.5\%$ as shown in bottom two rows.

\begin{table}[t]
\centering
\caption{Text-Caption CIDEr scores on the TextCaps dataset~\cite{sidorov2020textcaps}. The full result table can be found in the supplementary material.}
\vspace{-0.0in}
\begin{tabular}{ l c c }
    \hline
    Method & Val CIDEr & Test CIDEr\\
    \hline
    BUTD~\cite{anderson2018bottom} & 41.9 & 33.8 \\
    AoANet~\cite{huang2019attention} & 42.7 & 34.6 \\
    M4C~\cite{sidorov2020textcaps} & 89.6 & 81.0 \\
    MMA-SR~\cite{wang2020multimodal} & 98.0 & 88.0 \\
    CNMT~\cite{cnmt} & - & 93.03 \\
    M4C$^\dagger$~\cite{sidorov2020textcaps} & 99.89 & 93.36 \\
    \rowcolor{LightGray}
    TAP~(Ours) & 105.05 & 99.49 \\
    \rowcolor{LightGray}
    TAP$^{\dagger\dagger}$~(Ours) & \textbf{109.16} & \textbf{103.22}\\
    \hline
\end{tabular}
\vspace{-0.0in}
\label{table:textcaps}
\end{table}
\noindent\textbf{TextCaps.}
Table~\ref{table:textcaps} shows the CIDEr score on the TextCaps dataset~\cite{sidorov2020textcaps}. We report only the CIDEr score in the table and present the full table with other metrics in the supplementary material. We draw similar observations that with the same training data, ``TAP'' improves the CIDEr score of ``M4C$^\dagger$'' from $99.89$ to $105.05$. Large-scale pre-training ``TAP$^{\dagger\dagger}$'' further improves the CIDEr score to $109.16$.

\subsection{Ablation studies}
\label{sec:ablation}

\noindent\textbf{Pre-training tasks.}
We experiment with different pre-training tasks (MLM, ITM, RPP) as well as their variants. We conduct ablation studies on TextVQA with Microsoft-OCR and no extra data. We examine the effectiveness of scene-text language pre-training (MLM, ITM) and scene-text visual pre-training (RPP). We verify the importance of the extra scene-text token input $\mathbf{w^{ocr}}$ in MLM and ITM.

As shown in Table~\ref{table:obj}, the scene-text language pre-training in row $(d)$ and scene-text visual pre-training in row $(e)$ improve the non-TAP baseline (row $(b)$) from $44.50\%$ to $49.01\%$ and $46.42\%$, respectively. ``TAP'' performs all pre-training tasks and further improves the accuracy to $49.91\%$.

The extra scene text token input $\mathbf{w^{ocr}}$ is essential for TAP. Rows $(a$-$d)$ in Table~\ref{table:obj} show that neither extra $\mathbf{w^{ocr}}$ inputs (\cf rows $(a,b)$) nor pre-training (\cf rows $(b,c)$) alone lead to an improvement from the Non-TAP baseline (row $(b)$).
In contrast, TAP with the extra $\mathbf{w^{ocr}}$ input (row $(d)$) boosts the accuracy to $49.01\%$.
The bottom rows $(e,f)$ show the effectiveness of RPP. RPP with a single spatial relationship ``on'' improves the accuracy from $44.50\%$ to $46.42\%$ (\cf rows $(b,e)$). Combining RPP with MLM and ITM improves the accuracy from $49.01\%$ to $49.91\%$ (\cf rows $(d,f)$). Extending spatial relationship classes to $12$~\cite{yao2018exploring} leads to an improvement from $49.91\%$ to $50.17\%$.
\newcolumntype{P}[1]{>{\centering\arraybackslash}p{#1}}
\begin{table}[t]\fontsize{10}{11}\selectfont
\centering
\caption{Ablation studies on different pre-training tasks (MLM, ITM, RPP), and the variant of excluding the extra scene-text token input $\mathbf{w^{ocr}}$ in MLM and ITM. We highlight ``TAP'' by underline.}
\vspace{-0.0in}
\begin{tabular}{l | P{1.2cm} P{0.8cm} | c }
    \hline
     & +{\footnotesize MLM,ITM} & +RPP & Val Acc. \\
    \hline
    (a) Non-TAP w/o $\mathbf{w^{ocr}}$ & - & - & 44.48 \\
    (b) Non-TAP & - & - & 44.50 \\
    (c) + {\footnotesize MLM,ITM} w/o $\mathbf{w^{ocr}}$ & \checkmark & - & 44.63 \\ 
    (d) + {\footnotesize MLM,ITM} & \checkmark & - & 49.01 \\
    (e) + RPP & - & \checkmark & 46.42\eat{46.52/46.32} \\
    (f) TAP & \checkmark & \checkmark & \underline{49.91} \\
    \hline
\end{tabular}
\vspace{-0.1in}
\label{table:obj}
\end{table}

\begin{table}[t]
\centering
\caption{Ablation studies on pre-training with extra data. We use the listed data only in pre-training and then fine-tune the model with the TextVQA dataset only. $(3,4)$ and $(0,12)$ indicate the layer numbers of the text and multi-modal transformers, respectively. We highlight ``TAP'' and ``TAP$^{\dagger\dagger}$'' by underline and bold.}
\vspace{-0.0in}
\begin{tabular}{l | P{1.0cm} P{0.8cm} P{0.8cm} P{1.0cm} | P{0.7cm} P{0.7cm} }
    \hline
    & \multirow{2}{*}{\footnotesize{TextVQA}} & \multirow{2}{*}{\footnotesize{ST-VQA}} & \multirow{2}{*}{\footnotesize{TextCaps}} & \multirow{2}{*}{\footnotesize{OCR-CC}} & \multicolumn{2}{c}{\footnotesize{Val Acc.}} \\
    & & & & & \footnotesize{$(3,4)$} & \footnotesize{$(0,12)$} \\
    \hline
    (a) & \checkmark & - & - & - & \underline{49.91} & 48.78 \\
    (b) & \checkmark & \checkmark & - & - & 50.57 & 49.64 \\
    (c) & \checkmark & \checkmark & \checkmark & - & 51.86 & 50.13 \\
    (d) & - & - & - & \checkmark & 52.10 & 54.03 \\
    (e) & \checkmark & \checkmark & \checkmark & \checkmark & 52.90 & \textbf{54.71} \\
    \hline
\end{tabular}
\vspace{-0.0in}
\label{table:data}
\end{table}

\noindent\textbf{Pre-training with extra data}
Table~\ref{table:data} breaks down the benefits of adopting different sources of extra data. We conduct experiments on the TextVQA dataset with Microsoft-OCR. 
TAP enables the use of weak data with no answer annotations in the pre-training stage such like TextCaps and OCR-CC, in addition to the Text-VQA datasets. Compared with ``TAP'' with no extra data, pre-training with ST-VQA and TextCaps improves the accuracy from $49.91\%$ to $50.57\%$ and $51.86\%$ (\cf, rows $(a,b)$, rows $(b,c)$). The large-scale pre-training with OCR-CC (row $(d)$) achieves the accuracy of $52.10\%$. Including all data during pre-training (row $(e)$) further improves the accuracy to $52.90\%$.

Furthermore, we find that the extra data benefits the use of large models. The original architecture consists of a $3$-layer text-only transformer and a $4$-layer multi-modal transformer. We experiment with a $12$-layer multi-modal transformer with the same structure as \eat{BERT-base}BERT$_\text{BASE}$~\cite{devlin2018bert}. We initialize the model from BERT$_\text{BASE}$ and remove the separate text transformer. We represent the two architectures as $(3,4)$ and $(0,12)$ in Table~\ref{table:data}, where the numbers indicate the text and multi-modal transformer layer numbers. With extra transformer layers, the accuracy without extra data drops from $49.91\%$ to $48.78\%$ (row $(a)$), while the accuracy with extra data increases from $52.90\%$ to $54.71\%$ (row $(e)$).

\subsection{How does TAP help?}
\label{sec:analyses}
\begin{table}[t]
\centering
\caption{The coreference scores with and without TAP. Numbers represent the attention score between two semantically corresponded tokens, averaged across all such token pairs in TextVQA. Higher coreference scores imply a better aligned representation.}
\vspace{-0.0in}
\begin{tabular}{ l c c }
    \hline
    Coref Type & W/O TAP & With TAP \\
    \hline
    Text Word $\rightarrow$ Scene Text & 0.0477 & \textbf{0.3514} \\
    Scene Text $\rightarrow$ Text Word & 0.0473 & \textbf{0.5206} \\
    Visual Object $\rightarrow$ Scene Text & 0.0045 & \textbf{0.0130} \\
    Scene Text $\rightarrow$ Visual Object & 0.0337 & \textbf{0.0680} \\
    \hline
\end{tabular}
\vspace{-0.0in}
\label{table:scores}
\end{table}
\begin{figure*}[t]
\begin{center}
  \centerline{\includegraphics[width=17.5cm]{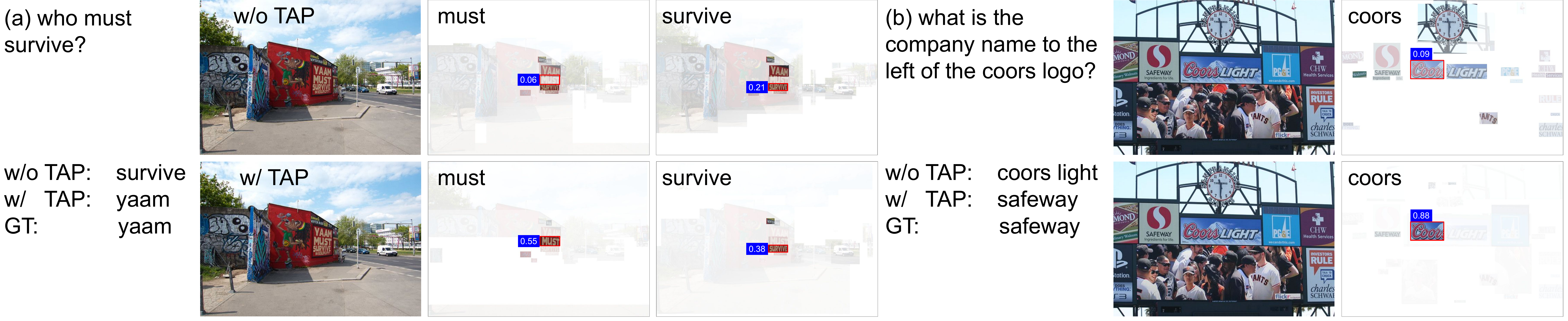}}
\end{center}
\vspace{-0.25in}
	\caption{Visualization of region attention scores with respect to each word in the question text $\mathbf{w}$, extracted from the multi-modal fusion transformers with (bottom row) and without (top row) TAP. The score by a region indicates its attention strength. TAP generates interpretable attentions on scene text-related question words like ``must'' and ``survive.''
}
\label{fig:attn}
\vspace{-0.1in}
\end{figure*}
In this section, we analyze how TAP helps Text-VQA/Text-Caption. We empirically show that with TAP, certain attention heads in the multi-modal transformer ground the scene text $\mathbf{v^{ocr}}$ to the semantically corresponded text word $\mathbf{w}$ or visual object $\mathbf{v^{obj}}$. By learning such latent alignments, TAP improves the aligned representation learning and thus helps Text-VQA/Text-Caption.

Recent VLP analyses~\cite{cao2020behind,li2020does} show that VLP~\cite{tan2019lxmert,chen2019uniter,li2019visualbert} learns the latent alignments between the semantically corresponded region-word or region-region pairs.
Specifically, certain attention heads in the transformer generate higher attention scores between such corresponded pairs.
The attention scores between corresponded pairs are also referred to as coreference scores~\cite{cao2020behind}. Similarly, we analyze the change in the coreference score of scene text-related pairs to better understand TAP.

There exist $(4$ layers$\times12$ heads$)=48$ attention scores between any two positions in our multi-modal transformer. Following VALUE~\cite{cao2020behind}, we define the coreference score as the maximum attention score among all $48$ heads between two semantically corresponded positions. 
A text word and a scene text region are corresponded if they refer to the same scene text token, \eg, the text word and scene text region ``coors'' in Figure~\ref{fig:attn}. We collect all corresponded pairs between the extended text input $\mathbf{w}$ and scene text regions $\mathbf{v^{ocr}}$ in the TextVQA dataset, and report the averaged score over all pairs. A scene text $\mathbf{v^{ocr}}$ and a visual object $\mathbf{v^{obj}}$ are corresponded if they share the spatial relationship ``on.''

As shown in Table~\ref{table:scores}, we analyze TAP by comparing the change in the coreference score before and after TAP, \ie, ``M4C$^\dagger$'' and ``TAP.'' The first two rows show that TAP improves the scene-text language coreference scores by seven times. The bottom two rows show that TAP increases the scene-text visual coreference scores by two times. These increases validate that TAP successfully learns the latent alignment and thus improves joint representation learning.

Furthermore, Figure~\ref{fig:attn} visualizes the attention score between a text word and all visual regions. 
Qualitatively, we observe a higher coreference score with TAP (bottom row) than the non-TAP baseline (top row). For example, in Figure~\ref{fig:attn} (a), TAP grounds the text word ``must'' and ``survive'' to the corresponded scene text regions.

\begin{figure*}[t]
\begin{center}
  \centerline{\includegraphics[width=16cm]{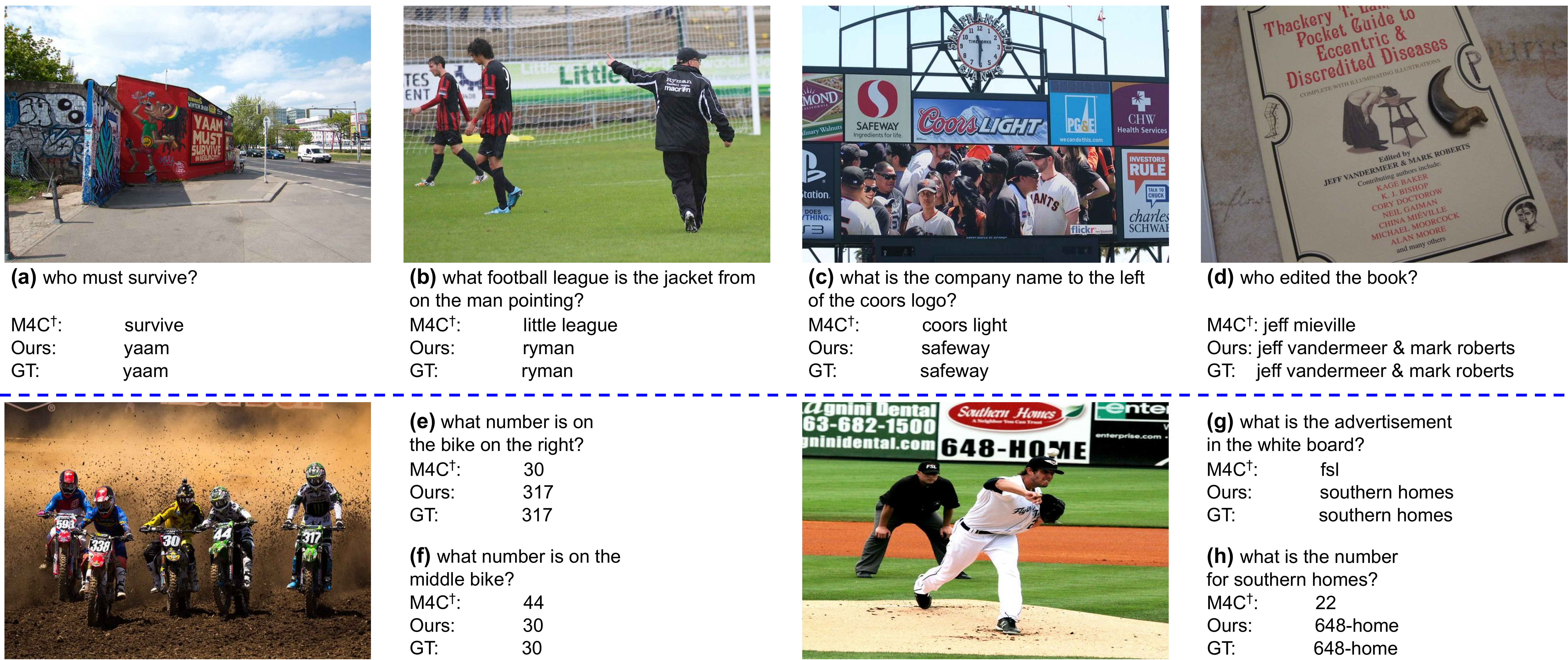}}
\end{center}
\vspace{-0.25in}
	\caption{Failure cases of the non-TAP baseline ``M4C$^\dagger$'' that can be corrected by ``TAP.''}
\label{fig:visu}
\vspace{-0.1in}
\end{figure*}
\subsection{Qualitative results}
\label{sec:visu}

Figure~\ref{fig:visu} shows representative failure cases of the non-TAP baseline ``M4C$^\dagger$'' that can be corrected by ``TAP.'' These cases show that TAP improves Text-VQA/Text-Caption by learning better aligned representations. 
\vspace{-6pt}
\begin{itemize} 
\setlength\itemsep{-3pt}
\item
TAP shows a good performance on challenging questions that require paraphrasing the scene text sentences. For example, in Figure~\ref{fig:visu} (a), the model answers ``who \textit{must survive}'' by the scene text ``yaam must survive'' in the image. The attention in Figure~\ref{fig:attn} further visualizes the latent region-word alignments.
\item
TAP also performs better on questions that refer to a scene text via an intermediate object. For example, in Figure~\ref{fig:visu} (b), the model grounds the object region ``the jacket on the man pointing'' and generates the correct answer ``ryman'' with the scene text ``ryman football league'' on the man's jacket.
\item
Figure~\ref{fig:visu} (c) shows an example that TAP correctly understands the relative spatial relationship in question. 
\item
Furthermore, TAP helps the model read a large piece of text. For example, in Figure~\ref{fig:visu} (d), the model correctly answers the question ``who edited the book'' by finding the editors' names ``jeff vandermeer \& mark roberts.'' We note that each word is detected as a separate scene text region, \eg, ``jeff,'' ``\&,'' \etc, which makes the answer sequence prediction non-trivia.
\end{itemize}
\vspace{-3pt}

The bottom row of Figure~\ref{fig:visu} shows examples of multiple questions on the same image. For example, (e,f)~(g,h) show that the model selects correct scene text regions as the answer based on the input questions. More qualitative results are included in the supplementary material.

%% file: conclusion.tex
\section{Conclusion}
We have presented Text-Aware Pre-training (TAP) that explicitly incorporates scene text in pre-training and effectively learns a better aligned multi-modality representation for Text-VQA/Text-Caption. With the identical framework and training data, TAP boosts the non-TAP baselines by $+5.4\%$ in absolute accuracy on the TextVQA challenge. Furthermore, we build a large-scale dataset named OCR-CC and further improve the TAP performance. TAP outperforms the state-of-the-art methods by large margins. Analyses show that TAP helps the aligned representation learning among text word, visual object, and scene text.

%% file: supply.tex
\twocolumn[{
\begin{center}
\Large 
\textbf{TAP: Text-Aware Pre-training for Text-VQA and Text-Caption}\\(Supplementary Material)
\par
\end{center}
\vspace{2em}
}]
\section{The OCR-CC Dataset}
\vspace{-0.2in}
\begin{figure}[h]
\begin{center}
  \centerline{\includegraphics[width=8.3cm]{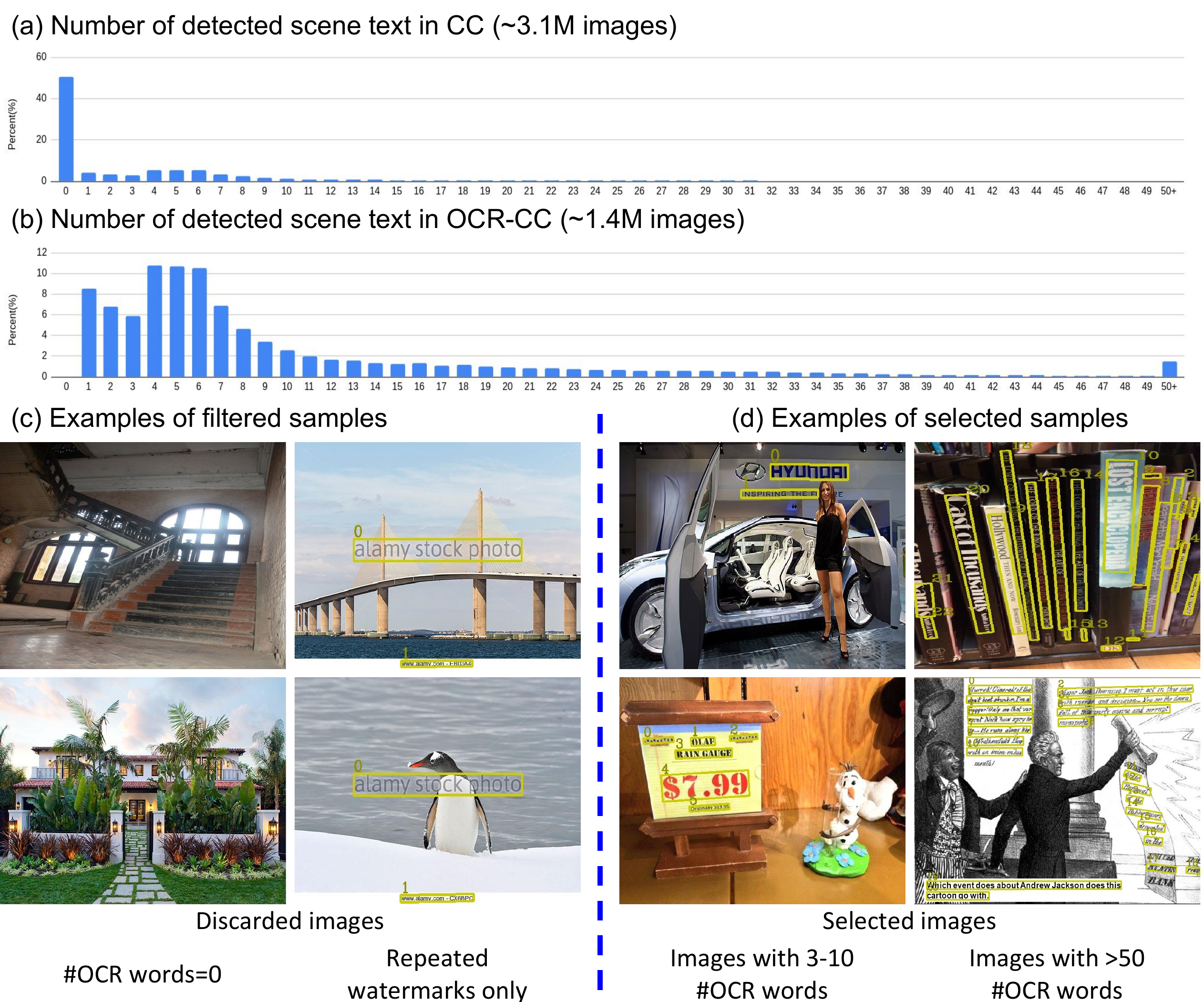}}
\end{center}
\vspace{-0.25in}
    \caption{\textbf{(a,b)} The distribution of the detected scene text number by Microsoft-OCR on the Conceptual Captioning (CC) dataset~\cite{sharma2018conceptual} and our OCR-CC dataset. \textbf{(c,d)} Representative examples of discarded and selected images. We draw the OCR box over multiple related words for visualization purposes. We note that each scene text region contains a single word, \eg, four words ``HYUNDAI,'' ``INSPIRING,'' ``THE,'' ``FL'' in the top left sub-figure of (d).}
\vspace{-0.1in}
\label{fig:ocrcc}
\end{figure}
In this section, we introduce the details of building the OCR-CC dataset based on the Conceptual Captioning (CC) dataset~\cite{sharma2018conceptual}. First, we run the Microsoft Azure OCR system on all CC images (around $3.1$ million). Then, we discard the images that don't have scene text (around half of the CC images) or have watermark ``text'' only (around $5\%$ of the CC images). These watermark ``text'' records the source image website/provider and are thus not related to the image content.  Figure~\ref{fig:ocrcc} (c) shows examples of the discarded images, which either have no detected scene text or have watermark ``text'' only. In the end, we select $1,367,170$ images from CC as the images in our OCR-CC dataset. We pair each selected image with a caption $\mathbf{w}$ for pre-training. The caption text $\mathbf{w}$ is the concatenation of the original image caption $\mathbf{w^q}$ in CC, the detected object labels $\mathbf{w^{obj}}$, and the detected scene text words $\mathbf{w^{ocr}}$. Figures~\ref{fig:ocrcc} (a,b) visualize the distribution of the scene text number in CC and our OCR-CC, respectively. Similar to the distribution on TextVQA~\cite{singh2019towards} and ST-VQA~\cite{biten2019scene}, the majority of images contains $3$-$10$ detected scene text regions, while a small portion of images has a large number of scene text regions. Figure~\ref{fig:ocrcc} (d) shows some representative selected images.

\section{TextCaps Results}
\begin{table}[t]
\centering
\caption{Results on the TextCaps~\cite{sidorov2020textcaps} validation set. B-4, M, R, S, C short for BLEU, METEOR, ROUGE\_L, SPICE, CIDEr, respectively. The oracle analyses are shown in the \textcolor{gray}{gray text} color.}
\vspace{-0.0in}
\begin{tabular}{ l c c c c c}
    \hline
    Method & B-4 & M & R & S & C\\
    \hline
    BUTD~\cite{anderson2018bottom} & 20.1 & 17.8 & 42.9 & 11.7 & 41.9 \\
    AoANet~\cite{huang2019attention} & 20.4 & 18.9 & 42.9 & 13.2 & 42.7 \\
    M4C~\cite{sidorov2020textcaps} & 23.3 & 22.0 & 46.2 & 15.6 & 89.6 \\
    MMA-SR~\cite{wang2020multimodal} & 24.6 & 23.0 & 47.3 & 16.2 & 98.0 \\
    M4C$^\dagger$~\cite{sidorov2020textcaps} & 24.3 & 22.9 & 47.3 & 16.5 & 99.9 \\
    \rowcolor{LightGray}
    TAP~(Ours) & 25.2 & 23.4 & 47.7 & 16.9 & 105.0 \\
    \rowcolor{LightGray}
    TAP$^{\dagger\dagger}$~(Ours) &\textbf{25.8} & \textbf{23.8} & \textbf{47.9} & \textbf{17.1} & \textbf{109.2}\\
    \hline
    M4C~(GT OCR)~\cite{sidorov2020textcaps} & \textcolor{gray}{26.0} & \textcolor{gray}{23.2} & \textcolor{gray}{47.8} & \textcolor{gray}{16.2} & \textcolor{gray}{104.3} \\
    \hline
\end{tabular}
\vspace{-0.0in}
\label{table:capsfull_val}
\end{table}
\begin{table}[t]
\centering
\caption{Results on the TextCaps~\cite{sidorov2020textcaps} test set.}
\vspace{-0.0in}
\begin{tabular}{ l c c c c c}
    \hline
    Method & B-4 & M & R & S & C\\
    \hline
    BUTD~\cite{anderson2018bottom} & 14.9 & 15.2 & 39.9 & 8.8 & 33.8 \\
    AoANet~\cite{huang2019attention} & 15.9 & 16.6 & 40.4 & 10.5 & 34.6 \\
    M4C~\cite{sidorov2020textcaps} & 18.9 & 19.8 & 43.2 & 12.8 & 81.0 \\
    CNMT\cite{cnmt} & 20.0 & 20.9 & 44.4 & 13.5 & 93.0 \\
    M4C$^\dagger$~\cite{sidorov2020textcaps} & 20.4 & 20.7 & 44.6 & 13.6 & 93.4 \\
    \rowcolor{LightGray}
    TAP~(Ours) & 21.5 & 21.7 & 45.4 & 14.5 & 99.5 \\
    \rowcolor{LightGray}
    TAP$^{\dagger\dagger}$~(Ours) &\textbf{21.9} & \textbf{21.8} & \textbf{45.6} & \textbf{14.6} & \textbf{103.2}\\
    \hline
    M4C~(GT OCR)~\cite{sidorov2020textcaps} & \textcolor{gray}{21.3} & \textcolor{gray}{21.1} & \textcolor{gray}{45.0} & \textcolor{gray}{13.5} & \textcolor{gray}{97.2} \\
    Human~\cite{sidorov2020textcaps} & \textcolor{gray}{24.4} & \textcolor{gray}{26.1} & \textcolor{gray}{47.0} & \textcolor{gray}{18.8} & \textcolor{gray}{125.5} \\
    \hline
\end{tabular}
\vspace{-0.0in}
\label{table:capsfull_test}
\end{table}

Tables~\ref{table:capsfull_val}, \ref{table:capsfull_test} present the full results on TextCaps~\cite{sidorov2020textcaps} to supplement the abstracted results in the main paper's Table~3. We draw similar conclusions from Tables~\ref{table:capsfull_val}, \ref{table:capsfull_test} as the ones in the main paper. Specifically, ``TAP'' significantly improves the non-TAP baseline ``M4C$^\dagger$'' in all metrics with the identical network architecture and training data. Our TAP approach also outperforms the previous state of the art~\cite{sidorov2020textcaps,wang2020multimodal,cnmt} by large margins. 

Furthermore, we compare TAP with the oracle numbers, as shown in the \textcolor{gray}{gray text} color at the bottom part of Tables~\ref{table:capsfull_val}, \ref{table:capsfull_test}. ``TAP'' outperforms the ``M4C~(GT OCR)'' that uses ground-truth scene text detection in training and inference. 
Meanwhile, there still exists a gap between ``TAP'' and human performance. We expect future studies focusing on captioning to further reduce the gap, \eg, with better decoding step pre-training designed especially for captioning.

\section{Hyper-parameters}
\label{sec:param}

We summarize the hyper-parameters used in the ``TAP'' and ``TAP$^{\dagger\dagger}$'' experiments. We conduct experiments based on the M4C~\cite{hu2020iterative,sidorov2020textcaps} and follow most of its hyper-parameter selections, as shown in Table~\ref{table:param}. We highlight the changed parameters in bold in the table.
\vspace{-6pt}
\begin{itemize} 
\setlength\itemsep{-3pt}
\item
First, the max length of the extended text input $\mathbf{w} = \left[\mathbf{w^q}, \mathbf{w^{obj}}, \mathbf{w^{ocr}}\right]$ is set to $20+100+100=220$. 
\item
Second, we increase the max length of scene text $\mathbf{v^{ocr}}$ from $50$ to $100$ when experimented with Microsoft-OCR.
Compared with Rosetta, Microsoft-OCR generates more detected scene text regions in each image. For example, in the TextVQA dataset, the mean and median of scene text numbers are $12.8$ and $8$ with Rosetta, and are $23.1$ and $12$ with Microsoft-OCR. With Rosetta, $3.5\%$ of images contain more than $50$ scene text regions detected, while the percentage is $14.3\%$ with Microsoft-OCR. To cover more detected scene text, we increase the max length of scene text $\mathbf{v^{ocr}}$ from $50$ to $100$ when experimented with Microsoft-OCR.
\item
In the experiment of ``pre-training without extra data'' (``TAP''), we follow the same learning rate step and maximum iteration settings as used in the fine-tuning. In pre-training with OCR-CC (``TAP$^{\dagger\dagger}$''), we pre-train the model for a maximum iteration of $480K$ and scale the learning rate steps linearly.
\end{itemize}
\begin{table}[t]
\centering
\caption{Hyper-parameters of the TAP experiments with and without OCR-CC pre-training, \ie, ``TAP$^{\dagger\dagger}$'' and ``TAP.'' We conduct the experiments based on M4C~\cite{hu2020iterative,sidorov2020textcaps} and highlight the changed parameters in bold. We detail these changes in Section~\ref{sec:param}.}
\vspace{-0.0in}
\begin{tabular}{ l c}
    \hline
    Hyper-parameter & Value \\
    \hline
    \textcolor{blue}{(a) General parameters} & \\
    max length of text word $\mathbf{w}$ & \textbf{220} \\
    max length of visual object $\mathbf{v^{obj}}$ & 100 \\
    max length of scene text $\mathbf{v^{ocr}}$ & \textbf{100} \\
    optimizer & Adam \\
    batch size & 128 \\
    base learning rate & 1e-4 \\
    warm-up learning rate factor & 0.2 \\
    warm-up iterations & 2000 \\
    max gradient L2-norm for clipping & 0.25 \\
    learning rate decay & 0.1 \\
    \hline
    \textcolor{blue}{(b) Pre-training parameters} & \\
    learning rate steps (``TAP,'' VQA) & 14K, 19K\\
    max iterations (``TAP,'' VQA) & 24K\\
    learning rate steps (``TAP,'' Caption) & 10K, 11K\\
    max iterations (``TAP,'' Caption) & 12K\\
    learning rate steps (``TAP$^{\dagger\dagger}$'') & \textbf{280K, 380K}\\
    max iterations (``TAP$^{\dagger\dagger}$'') & \textbf{480K}\\
    \hline
    {\small\textcolor{blue}{(c) Text-VQA Fine-tuning (TextVQA, ST-VQA)}} & \\
    max length of decoding step & 12 \\
    learning rate steps & 14K, 19K\\
    max iterations & 24K\\
    \hline
    \textcolor{blue}{(d) Text-Caption Fine-tuning (TextCaps)} & \\
    max length of decoding step & 30 \\
    learning rate steps & 10K, 11K\\
    max iterations & 12K\\
    \hline
\end{tabular}
\vspace{-0.0in}
\label{table:param}
\end{table}

\section{Pre-train + Fine-tune \vs Joint-train}
Results in the main paper's Section~4.3 show that TAP works well even without extra data. We hypothesize that we can view TAP as a multi-task learning framework, and obtain similar improvement by using the pre-training tasks (MLM, ITM, RPP) as the auxiliary training loss. Therefore, we explore an alternative training pipeline named ``joint train,'' where the pre-training tasks are used as the auxiliary losses together with the main answer/caption loss. Because MLM and ITM tasks require ``polluting'' the input sequence, we randomly select $50\%$ of the samples in a batch to compute the pre-training loss and keep the remaining $50\%$ unchanged for the answer/caption loss.

Studies show that these two training pipelines can achieve similar performances, \ie, $49.91\%$ for ``pre-train + fine-tune'' and $49.46\%$ for ``joint train'' on TextVQA. Both methods significantly outperform the non-TAP baseline ($44.50\%$). For ``joint train,'' we train the framework for $120$K iterations. Compared with ``joint train,'' one advantage of the ``pre-train + fine-tune'' pipeline in the main paper is that the extra weak data with no answer/caption annotations can be more easily used. 

The effectiveness of different TAP pipelines implies the potential of improving other multi-modal tasks by incorporating pre-training tasks. Specifically, the pre-training tasks can be used either in the ``joint-train'' approach to best preserve the main task's training pipeline, or in the ``pre-train + fine-tune'' approach to benefit from the large-scale weak pre-training data.

\section{Qualitative Results}
\begin{figure*}[t]
\begin{center}
   \centerline{\includegraphics[width=17cm]{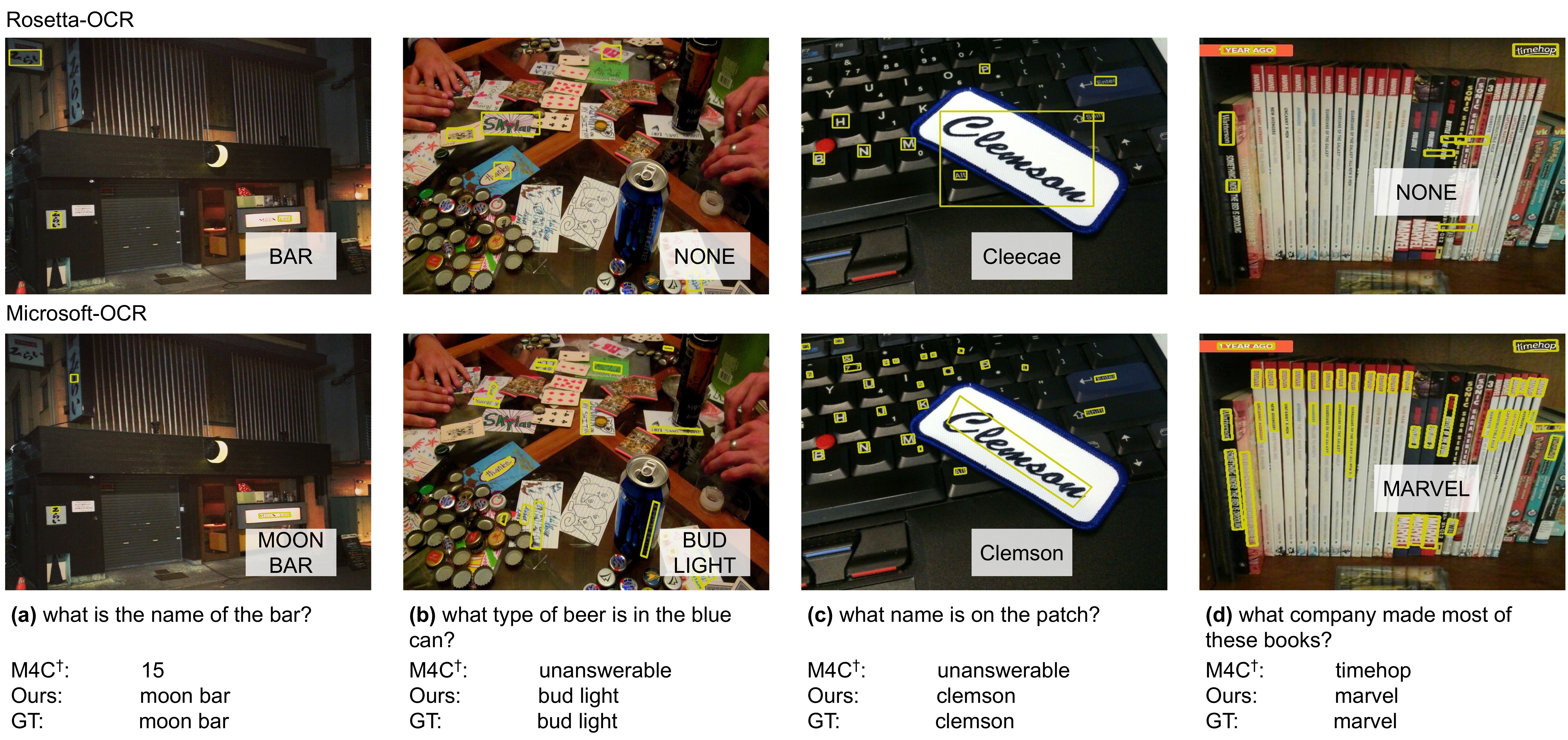}}
\end{center}
\vspace{-0.2in}
	\caption{Failure cases that can be corrected by scene text detection. The top and bottom rows visualize the detected scene text by Rosetta-OCR and Microsoft-OCR, respectively. We draw adjacent words into the same box for visualization purposes and highlight the key scene text regions for the question, \eg, ``moon bar,'' ``bud light,'' ``clemson,'' and ``marvel.''}
\label{fig:ocr}
\end{figure*}

\begin{figure*}[t]
\begin{center}
   \centerline{\includegraphics[width=17cm]{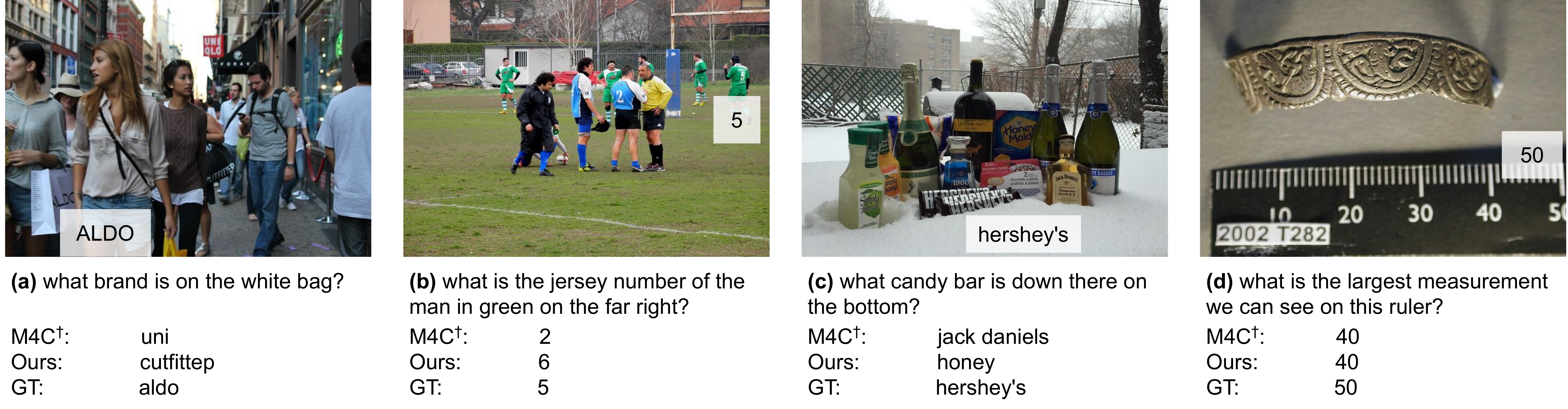}}
\end{center}
\vspace{-0.15in}
	\caption{Representative failure cases of ``TAP.'' We highlight the key scene text regions for each question.}
\label{fig:failure}
\vspace{2.9in}
\end{figure*}

In this section, we present additional qualitative examples. Figure~\ref{fig:ocr} shows the failure cases that can be corrected by OCR detection. Figure~\ref{fig:failure} presents the failure cases of our method. ``TAP'' occasionally fails on samples that require complex reasoning (Figures~\ref{fig:failure} (a,b)) or have incorrect scene text detection (Figures~\ref{fig:failure} (c,d)). For example, in Figure~\ref{fig:failure} (a), TAP selects the scene text ``cutfittep'' on the black bag as the answer, instead of the correct scene text ``aldo'' on the referred white bag.

